# Representation Heterogeneity *

Fausto Giunchiglia, Mayukh Bagchi

*Department of Information Engineering and Computer Science, University of Trento, I-38123 Povo, Trento, Italy*

**Abstract**
Semantic Heterogeneity is conventionally understood as the *existence of variance* in the representation of a target reality when modelled, by independent parties, in different databases, schemas and/ or data. We argue that the mere encoding of variance, while being *necessary*, is *not sufficient* enough to deal with the problem of representational heterogeneity, given that it is also necessary to encode the unifying basis on which such variance is manifested. To that end, this paper introduces a notion of *Representation Heterogeneity* in terms of the co-occurrent notions of *Representation Unity* and *Representation Diversity*. We have representation unity when two heterogeneous representations model the same target reality, representation diversity otherwise. In turn, this paper also highlights how these two notions get instantiated across the two layers of any representation, i.e., *Language* and *Knowledge*.

## 1. Introduction

The phenomenon of *Semantic Heterogeneity* is conventionally understood as the *existence of variance* in the representation of the same target reality when computationally modelled by independent parties in database schemas or data sets [1]. The principal ramifications of semantic heterogeneity in data management include *representational incompleteness* and *inconsistency* with the consequent *loss of semantic interoperability* [2]. This is a widely studied problem in increasingly emergent application scenarios like multilingual data integration (see, for instance, [3, 4]) and many *partial solutions* at the schema and data level have been proposed (see, for instance, [5, 6]). However, so far, there is no unifying model of *why* and *how* semantic heterogeneity manifests itself and, even less, a general solution.

We prefer to talk of *Representation Heterogeneity*, rather than of *Semantic Heterogeneity*, to emphasize the fact that heterogeneity is an intrinsic property of any representation [7], wherein different observers encode different representations of the same target reality depending on the local context [8, 9]. Representation heterogeneity is, in turn, rooted in the more general omnipresent phenomenon of *World Heterogeneity*. Thus, for instance, there is a need of determining whether two different (occurrences of) musical instruments are actually the same instrument. In this perspective, we define the problem of representation heterogeneity as follows. Given that (i) there are no two identical occurrences of reality, not even of the same reality, and that

*This research has received funding from the *"DELPhi - DiscovEring Life Patterns"* project funded by the MIUR (PRIN) 2017.
*1st International Workshop on Formal Models of Knowledge Diversity, Joint Ontology WOrkshops (JOWO), Sweden*
✉ fausto.giunchiglia@unitn.it (F. Giunchiglia); mayukh.bagchi@unitn.it (M. Bagchi)
 0000-0002-5903-6150 (F. Giunchiglia); 0000-0002-2946-5018 (M. Bagchi)

(ii) there are no two identical representations, of even of the same occurrence of reality, then (iii) how can we establish whether *two heterogeneous representations actually represent the same reality?*

We argue that the current understanding of semantic heterogeneity as the *'existence of variance'*, while being crucially *necessary*, is not *sufficient*. *There can be no variance without a prior notion of a unifying reference taken as the basis for computing the variance itself.* To that end, we propose to ground the notion of representation heterogeneity in that of world heterogeneity which, in turn, we model as the *co-occurrence* of *(World) Unity* and *(World) Diversity*. *Unity*[1] models the ability of recognizing two different real world phenomena as *different occurrences* of the same target reality, and, given *Unity*, *Diversity* models the ability of recognizing the existence of differences among them. In turn, this allows to disambiguate the (also omnipresent) phenomenon of representation heterogeneity into the two distinct phenomena of *Representation Unity* which models the fact that two representations represent the same target reality, and given *Representation Unity*, the notion of *Representation Diversity* models the differences between the representations. Finally, we model these two notions it in terms of the co-occurrence of *Unity* and *Diversity* into two distinct ordered layers, i.e., *Language* and *Knowledge*. The *Language* layer comprises the heterogeneity arising in the conceptual and lexical-semantic level. Instead, the *Knowledge* layer comprises of the heterogeneity arising in the schema and data level.

This paper is organized as follows. Section 2 introduces the notions of world and representation heterogeneity. Section 3 unwinds the notion of representation heterogeneity into the notions of *Language* and *Knowledge* unity and diversity. Section 4 concludes the paper with a short comparison with Brachman and Guarino's proposed layering of representations.

## 2. World heterogeneity

Consider the motivating example (see Table 1) of the two datasets encoding information about the same target reality - a musical instrument identified as `2290SDC50`. The first dataset is a record in a musical instruments catalog from Europe capturing some geophysical details of the aforementioned instrument such as production, collection, width etc. The second dataset, instead, is a record from the instrument's host museum in India encoding details such as its company and width. There are at least four levels which complicate the *representation* of the entity (with label) `2290SDC50`. First, the fact that the same musical instrument is conceptualized differently in the two datasets, *viz.* chordophone and stringed instrument respectively. Second, but non-trivially, the fact that the first dataset is in English whereas the second dataset is in Hindi. Third, the observation that for modelling the same entity, each dataset employ a different set of properties thus, essentially, leading to two different descriptions. Finally, the fact that even for a common property such as the width, the values recorded are different due to the adherence to different units of measurement.

This example is a direct instantiation of the phenomenon of *Representation Heterogeneity*, which is conventionally understood as the *existence of variance* in the representation, interpretation and resulting meaning. Representation heterogeneity is pervasive, wherein, even for the same target reality, different observers encode different representations depending on the local context, purpose, focus or other factors. In turn, representation heterogeneity is rooted in the *unavoidable* phenomenon of *World Heterogeneity*. Genetic diversity allows species to adapt to changes in the environment, production diversity allows economies to adapt to changes in market dynamics, and social and cultural diversity fuel progress in the society. Heterogeneity is the key distinguishing feature of life, there will never be, e.g., two identical places or two identical individuals. Still, despite this, we are able to determine whether two heterogeneous occurrences of reality are actually occurrences of the same reality. Thus for instance, we can determine whether (or not) two different (occurrences of) objects are two instances of a musical instrument or whether (or not) two different (occurrences of) musical instruments are two occurrences of the same instrument. We formalize the intuitions above as follows.

---

[1]The notion of *Unity* is unrelated to its namesake in OntoClean [10]

**Table 1**
Two heterogeneous datasets encoding information about musical instruments.

| Chordophone | | | | |
|---|---|---|---|---|
| List No. | Production | Collection | Width | Date |
| 2290SDC50 | India | SDC Museum | 162 | 26-06-1950 |

| तंतु वाद्य (Stringed Instrument) | | |
|---|---|---|
| सूची संख्या (Inventory No.) | कंपनी (Company) | चौड़ाई (Width) |
| २२९०एसडीसी५० (2290SDC50) | कला भवन (Kala Bhawan) | ६.३८ (6.38) |

*World Heterogeneity:* there are no two identical occurrences of the same or different realities;

*Representation Heterogeneity:* there are no two identical representations of the same or different occurrences of reality.

Notice how the latter is an instance of the former and, as such, it is also unavoidable, as it is the consequent *(non-)generality of any representation* [11]. It is *impossible to construct a representation capable of capturing the infinite richness of the real world* and also the infinite ways, provided by language, to describe the world itself. Thus, on one hand, for any chosen representation, there will always be an aspect of the world which is not captured and, on the other hand, there will always be an alternative way to represent the same aspect of the world.

Based on these premises, it should be evident that the understanding of semantic heterogeneity as the *'existence of variance'*, while being crucially *necessary*, is not *sufficient* to characterize the heterogeneity of representations, and even less to suggest a way to handle it. The crucial observation is that, being everything (and every representation) different from everything else (and any other representation), *a proper notion of variance can only be given based on a prior notion of a unifying reference taken as the basis for computing the variance itself*. And, once defined the unifying reference, we need also, as a second step, dependent on the previous, to make precise the basis on which we compute what is different.

## 3. Representation heterogeneity

We model the representation heterogeneity in terms of the *co-occurrence* of the two notions of *Representation Unity* and *Representation Diversity*. *Representation Unity* models the fact that two different representations actually *represent* the same target reality, e.g., two encounters with the same chordophone `2290SDC50`, or two encounters with a *musical instrument*. Given a *Representation Unity*, namely for any two representations for which it has been identified the reason why there is Unity, *Representation Diversity* models the ability of recognizing their mutual differences. Thus, for instance, we can recognize that two musical instruments (unity) are a *chordophone* and an *aerophone* (diversity), or the fact that two musical instruments have different width. Thus, while establishing *Representation Unity* allows to *determine the space of what exists* across multiple perceptions of the same target reality, establishing *Representation Diversity* allows to *determine the space of variations*, e.g., the properties, of any target reality which was decided to exist. Based on this, we propose the following *solution to the problem of representation heterogeneity*:

> For any two representations, given that:
>> *(representation heterogeneity):* there are no two identical representations of reality,
>
> decide whether:
>> *the two representations represent the same reality.*

If this is the case then we say that we have the unity of representations, otherwise we say that we have representation diversity. But how to establish this fact? We do this by recursively reducing the problem of representation heterogeneity to the problems of language and knowledge heterogeneity and, in turn, to the co-occurrence of *Unity* and *Diversity* in the *Language* and the *Knowledge* layers. Let us briefly highlight how this works in practice.

In the *Language* layer, the *need* for co-occurrence of *Language Unity* and *Language Diversity* are primarily due to the different conceptual hierarchies into which (the same) target realities are hierarchically modelled, in terms of *Genus* and *Differentia* [12, 13]. Thus, for instance, *Koto*, *Dulcimer* and *Guitar* are different while being *string instruments*. *Language Unity* models the ability of recognizing two different concepts as *different occurrences* of the same common concept. Thus for instance, in the above example, we establish language unity by establishing that we have musical instruments. *Language Diversity* models the ability of recognizing the differences between them, in the example above, that we have two different musical experiments. So, we always have both language unity and language diversity. The one which is selected depends on the level of abstraction at which we are thinking. Are we looking for two musical instruments, and we do not care which one, or are we looking for a specific one, e.g., a Guitar?

A similar situation happens in the *Knowledge* layer. If in the language layer we need to define the objects we are looking for, in the knowledge layer we need to clarify the specific properties (of such objects) we are interested in. *Knowledge Unity* models the ability of recognizing two different entities as *different occurrences* of the same common entity, while, for any occurrence of *Knowledge Unity*, *Knowledge Diversity* models the ability of recognizing the differences between the entities. Thus, for instance, once we have decided at the language level that we are interested in Guitars, we may establish that we are interested in guitars of a certain form, or of a certain color. The selected set of properties determines which unity and diversity we are looking for.

The process by which we decide on representation heterogeneity will lead to different results depending on the specifics of what we are looking for, as it is the case in our everyday life. We formalize this in the following notion

*Language Heterogeneity:* whether there is language unity or language diversity depends on the selected level of abstraction in the Genus-Differentia hierarchy;

*Knowledge Heterogeneity:* given a certain language unity, whether there is knowledge unity or knowledge diversity depends on the properties selected as reference.

This leads to the following refined *solution to the problem of representation heterogeneity*:

> For any two representations, given that:
>> *(representation heterogeneity):* there are no two identical representations of reality,
>
> decide whether:
>> *the two representations represent the same reality,*
>
> based on:
>> *the selected language level of abstraction and follow-up selected knowledge level properties.*

In other words, a general solution to the problem of semantic/ representation heterogeneity must be parametric on the local *purpose*, with the prupose being characterized in terms of two precise choices at the language and knowledge level.

## 4. Conclusion - on layering representations

| Level | Primitive concepts... | Main feature | Interpretation |
| --- | --- | --- | --- |
| Logical | are predicates | Formalisation | Arbitrary |
| Epistemological | are structuring primitives | Structure | Arbitrary |
| *Ontological* | *satisfy meaning postulates* | *Meaning* | *Constrained* |
| Conceptual | are cognitive primitives | Conceptualisation | Subjective |
| Linguistic | are linguistic primitives | Language | Subjective |

**Figure 1:** Knowledge Representation Levels - Guarino (taken from [14]).

Our solution to the problem of semantic heterogeneity is based on a *stratification* of representations where, starting from the heterogeneity of the world, we articulate the unity and diversity of two representations in terms of a set of choices made at the language and knowledge level.

This is not the first time a stratification of representation has been provided. Most noticeable is the model proposed first by Brachman [15] and later modified by Guarino [14, 16] (see Figure 1). Withe respect to Brachman's proposal, Guarino emphasized that the focus of the epistemological level was on structuring and *formal reasoning* and not on *formal representation* of concepts which remained arbitrary and *neutral* as concerns ontological commitment. He argued against this very *ontological neutrality* and advocated that a rigorous *ontological analysis* can greatly *"improve the quality of the knowledge engineering process"* [14]. To that end, he proposed a new knowledge representation level termed as the *ontological level* positioned between epistemological and conceptual level (see Figure 1). The *ontological level* was intended as the *"level of meaning"* and offered primitives which *" satisfy formal meaning postulates ... restrict[ing] the interpretation of a logical theory on the basis of formal ontology, intended as a theory of a priori distinctions"* [14]. As part of the ontological level, several distinctions (in the form of *metaproperties*, e.g., *sortal*, *rigidity*) were proposed (see [16] for full details).

The stratification of representation proposed here is orthogonal to the one proposed by Brachman and Guarino. Their work was focused on the knowledge engineering process by which one would generate a certain model of reality. Our work is focused on how to solve the problem of semantic heterogeneity. The solution we propose is a stratified model of how we represent the world and of how we deal with semantic heterogeneity. Quoting from [17], representation heterogeneity should be seen a *"feature which must be maintained and exploited"* and not a *"defect that must be absorbed"*, as it is the means by which we cope with the complexity of the world heterogeneity. The work so far, as partially cited in this paper, make us hopeful.